\documentclass[sigconf]{acmart}

\usepackage{color}
\usepackage{multirow}
\usepackage{pifont}
\newcommand{\cmark}{\ding{51}}%
\newcommand{\xmark}{\ding{55}}%

\newcommand{\heading}[1]{\noindent\textbf{#1}}

\AtBeginDocument{%
  \providecommand\BibTeX{{%
    \normalfont B\kern-0.5em{\scshape i\kern-0.25em b}\kern-0.8em\TeX}}}
\renewcommand\footnotetextcopyrightpermission[1]{}

\begin{document}
\fancyhead{}

\title{You Only Align Once: Bidirectional Interaction for Spatial-Temporal Video Super-Resolution}

\author{Mengshun Hu} \author{Kui Jiang}\authornote{Equal contribution} \author{Zhixiang Nie} \author{Zheng Wang}\authornote{Corresponding author}
\affiliation{%
 \institution{National Engineering Research Center for Multimedia Software, Hubei Key Laboratory of Multimedia and Network Communication Engineering, School of Computer Science, Wuhan University \country{}}
}

\begin{abstract}
Spatial-Temporal Video Super-Resolution (ST-VSR) technology generates high-quality videos with higher resolution and higher frame rates. Existing advanced methods accomplish ST-VSR tasks through the association of Spatial and Temporal video super-resolution (S-VSR and T-VSR). These methods require two alignments and fusions in S-VSR and T-VSR, which is obviously redundant and fails to sufficiently explore the information flow of consecutive spatial LR frames. Although bidirectional learning (future-to-past and past-to-future) was introduced to cover all input frames, the direct fusion of final predictions fails to sufficiently exploit intrinsic correlations of bidirectional motion learning and spatial information from all frames. We propose an effective yet efficient recurrent network with bidirectional interaction for ST-VSR, where only one alignment and fusion is needed. Specifically, it first performs backward inference from future to past, and then follows forward inference to super-resolve intermediate frames. The backward and forward inferences are assigned to learn structures and details to simplify the learning task with joint optimizations. Furthermore, a Hybrid Fusion Module (HFM) is designed to aggregate and distill information to refine spatial information and reconstruct high-quality video frames. Extensive experiments on two public datasets demonstrate that our method outperforms state-of-the-art methods in efficiency, and reduces calculation cost by about \textcolor{blue}{22\%}. 
\end{abstract}

\begin{teaserfigure}
  \includegraphics[width=1.0\textwidth]{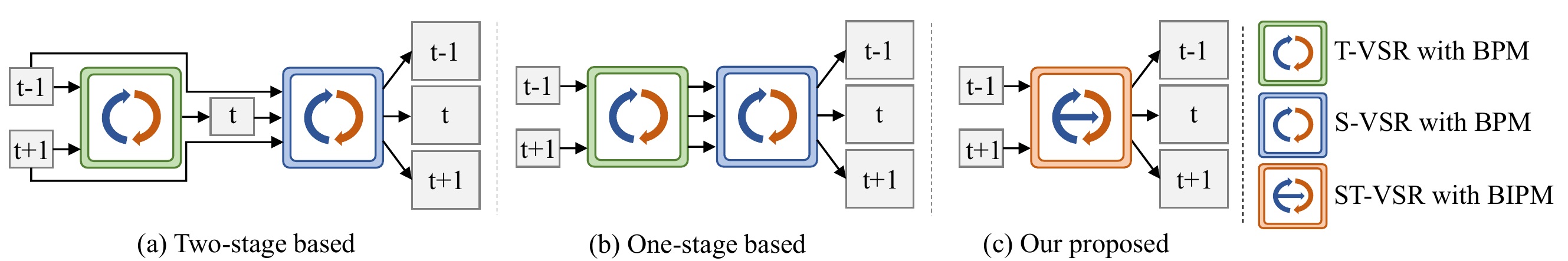}
  \caption{(a): Two-stage based methods: They perform ST-VSR by independently and sequentially using advanced S-VSR and T-VSR without merging common pipelines (\textit{i.e.} feature extraction, alignment, fusion and reconstruction), both involving bidirectional propagation module (BPM) for alignment and fusion on image space. (b) One-stage based methods: They unify S-VSR and T-VSR into a single stage for ST-VSR with sharing feature extraction and reconstruction network, both involving bidirectional propagation module (BPM) for alignment and fusion on feature space. (c) Our proposed method: We can accurately and fast super-resolve videos by efficiently merging common pipelines from S-VSR and T-VSR, only involving bidirectional interactive propagation module (BIPM) for once alignment and fusion on feature space.}
  \label{fig:teaser}
\end{teaserfigure}

\maketitle

\section{Introduction}
Spatial-temporal video super-resolution (ST-VSR) refers to the task of generating the high-resolution (HR) and high-frame-rate (HFR) photo-realistic video sequences from the given low-resolution (LR) and low-frame-rate (LFR) input. This task has drawn increasing attention and become a research hotspot in the field of multimedia~\cite{haris2020space,xiang2021zooming,xu2021temporal,you2022megan,zhou2021video}, because of the broad range of applications such as movie production~\cite{kim2020fisr}, high-definition television upgrades~\cite{kang2020deep} and video compression~\cite{xiang2021zooming}, \emph{etc}.

1) \textit{Two-stage based}: To tackle ST-VSR, quite intuitively, a direct combination of temporal video super-resolution (T-VSR)~\cite{bao2019depth} and spatial video super-resolution~\cite{jiang2020hierarchical} (S-VSR)~\cite{bao2019depth} can cope with some simple cases with small motions and scene changes. Sequential T-VSR and S-VSR barely considers the intrinsic relationship between these two tasks when generating HR and HFR videos, where rich spatial information can help temporal prediction and sequential information is crucial for texture inference (see Figure~\ref{fig:teaser}(a)). Therefore, this sort of two-stage based methods are far from producing satisfactory reconstruction results, especially for large motion and magnification factors. Moreover, separate T-VSR and S-VSR require repetitive operations such as feature extraction, alignment, fusion, and reconstruction, which is structurally redundant and inefficient.

2) \textit{One-stage based}: By contrast, another sort of methods tackle ST-VSR on feature space~\cite{xiang2020zooming,xiang2021zooming,xu2021temporal,you2022megan} by integrating T-VSR and S-VSR into a unified framework for joint optimization with shareable feature extraction and reconstruction modules (see Figure~\ref{fig:teaser} (b)). Although spatial-temporal correlation learning is improved with a compact ST-VSR framework, these methods still require separate alignment and fusion operations for spatial and temporal information representations. The motions with bidirectional propagation module (BPM) in T-VSR and S-VSR are independent and lack interactions. The reconstruction of the current frame does not sufficiently explore spatial-temporal correlations from past, current and future frames, and such long-term relationships are critical for large motion estimation.

We propose to further compact ST-VSR framework and sufficiently explore spatial-temporal correlations by learning the interaction of bidirectional inference. The proposed framework introduces a novel bidirectional inferences scheme and a Bidirectional Interactive Propagation Module (BIPM) to implicitly align and mine spatial-temporal information. In BIPM, two Recurrent Cells (RCs) are equipped to enable bidirectional interaction so that the current representation can absorb knowledge from the past, current and future frames. In this way, our proposed framework only requires one alignment operation to capture the spatial-temporal correlations from all frames, dubbed \underline{Y}ou \underline{O}nly ali\underline{G}n \underline{O}nce (YOGO). 

In addition, since spatial information may gradually vanish during the propagation process, instead of directly reconstructing HR frame from the output of two recurrent cells~\cite{xiang2020zooming,xu2021temporal,chan2021basicvsr}, we propose a Hybrid Fusion Module (HFM) to further progressively refine spatial information via the outputs of two recurrent cells. 

Our contributions are summarized as follows:
\begin{itemize}
\item We propose a novel yet high-efficiency framework for ST-VSR, namely YOGO. In YOGO, T-VSR and S-VSR are integrated into a unified network to promote the compact.
\item We devise a bidirectional interactive propagation scheme to explore spatial-temporal correlations, where past, current and future knowledge from all frames are aggregated by updating the hidden states.
\item We conduct extensive experiments to compare our YOGO on ST-VSR, which demonstrates our method performs well against the state-of-the-art ST-VSR methods in efficiency.
\end{itemize}

\begin{figure*}[t]
     \centering
     \includegraphics[width=1.0\textwidth]{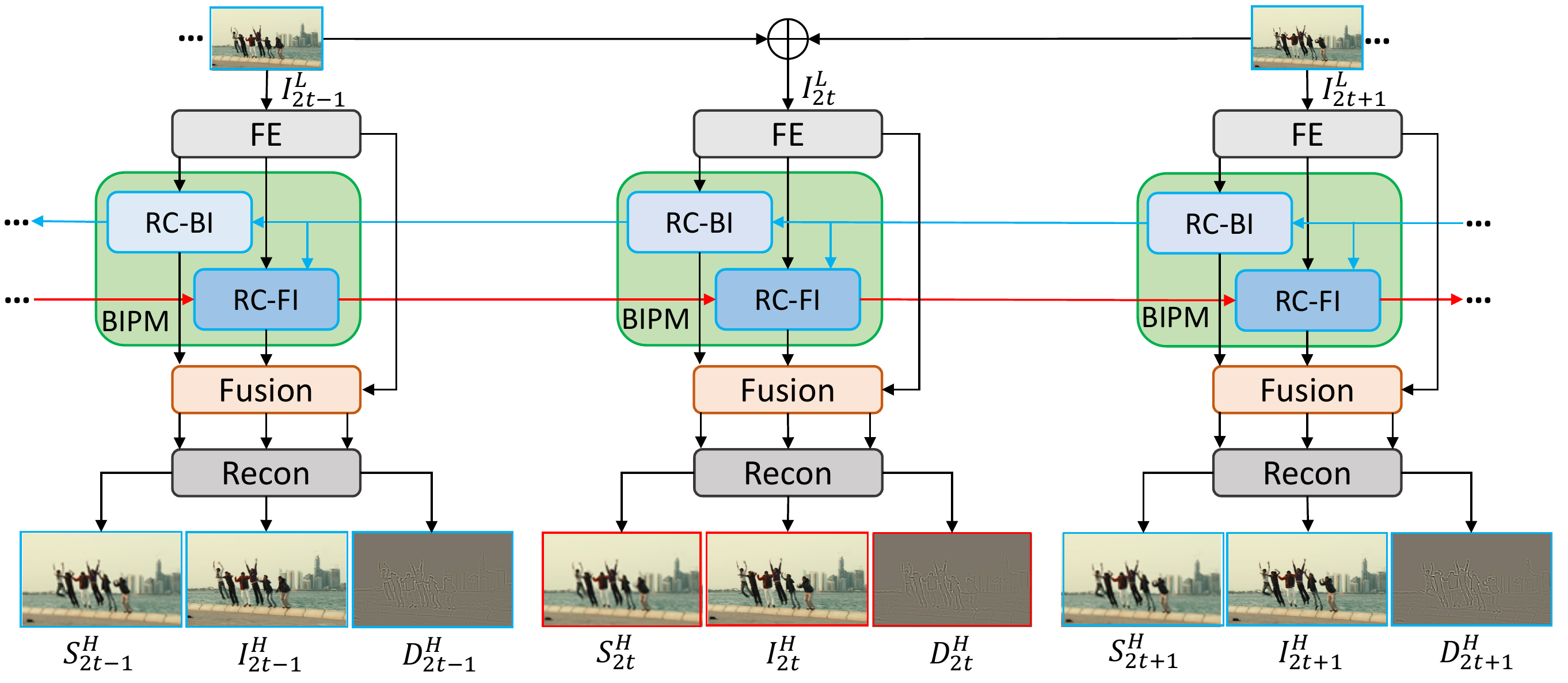}
     %\vspace{-2mm}
     \caption{\textbf{Architecture of the proposed YOGO.} Given multiple low-resolution (LR) and low-frame-rate (LFR) input frames, we firstly extract representations from input frames by feature extractor (FE). We then adopt a bidirectional interactive propagation module (BIPM) containing two recurrent cells (\textit{e.g.} recurrent cell with backward inference (RC-BI) and recurrent cell with forward inference (RC-FI)) to  implicitly align corresponding hidden states. Then, we assign backward and forward inferences to learn structures and details components from temporal information, and further progressively aggregate and distill structures and details components for spatial reconstruction via a hybrid fusion Module (HFM) in the Fusion process. Finally, we employ the reconstruction module (Recon) to output the high-resolution (HR) (x4)  and high-frame-rate (HFR) (x2) videos, structures and details components.}
     %\vspace{-2mm}
     \label{framework}
\end{figure*}

\section{Related Work}
%Our work is mainly related to deep learning based methods for
%three video super-resolution: spatial video super-resolution (S-VSR), temporal video super-resolution (T-VSR) and spatial-temporal video super-resolution (ST-VSR). 

\subsection{Spatial Video Super-resolution}
S-VSR aims to super-resolve LR videos to HR videos by sufficiently utilizing temporal information. Thus, the key to this task lies in making full use of temporal correlations among multiple frames. The existing S-VSR method can be mainly divided into two frameworks: sliding-window and recurrent frameworks. The former firstly conducts motion estimations between low-resolution frames from a sliding window, and then perform spatial alignments based on predicted motions and fusion for HR reconstruction ~\cite{caballero2017real, tao2017detail,wang2019edvr,bao2019memc,li2020mucan}. %For example, Xue \textit{et al.} adopt explicit motion estimation (\emph{e.g.}, optical flow) to align adjacent frames within the sliding window, then fusion all aligned frames and current frame for HR reconstruction via image processing module~\cite{xue2019video}. Tian \textit{et al.} apply deformable convolution~\cite{dai2017deformable,zhu2019deformable} for implicit motion estimation to replace the optical flow for further optimization~\cite{tian2020tdan}. 
However, these methods are time-consuming and each input frame is processed and aligned multiple times. Moreover, Since they cannot build long-range temporal correlations from input videos, they tend to generate temporally consistent results.

Unlike S-VSR methods based on sliding window framework, due to the recurrent property, S-VSR methods based on recurrent framework~\cite{huang2017video,chan2021basicvsr,chan2021basicvsr++,sajjadi2018frame,haris2019recurrent,li2021comisr} are able to explore and utilize long-range temporal correlations without multiple alignments for each frame. For example, RSDN~\cite{isobe2020video} proposes a recurrent dual-branch network to learn the structures and details of frames for S-VSR. But it fails to leverage the information from subsequent LR frames while conducting single-direction recurrent propagation. To alleviate this issue, Yi \textit{et al.} design a novel recurrent network with bidirectional coupled propagation for ST-VSR, which can make full use of long-range temporal correlations by implicitly aligning and fusing the past, current, future information from video sequences~\cite{yi2021omniscient}.   

\subsection{Temporal Video Super-resolution}
T-VSR (\emph{i.e.}, video frame interpolation) aims to generate non-existent intermediate frames between consecutive input frames. Thus, the key to this task lies in finding pixel correspondences between consecutive frames. The common methods for T-VSR are mainly divided into two categories: flow-based methods~\cite{bao2019depth,jiang2018super,hu2020motion,park2020bmbc,park2021asymmetric,sim2021xvfi} and kernel-based methods~\cite{cheng2021multiple,niklaus2017videob,shi2021video}. flow-based methods~\cite{gui2020featureflow} mainly consist of the following steps: feature extraction, forward and backward alignment, aligned intermediate representations fusion. intermediate frame reconstruction. To further refine reconstructed results, additional contextual information~\cite{bao2019memc,niklaus2018context,hu2021capturing,niklaus2020softmax} is introduced to a post-processing module to predict residual information for compensation. However, These methods rely heavily on optical flow estimation accuracy. As for kernel-based methods, some methods estimate dynamic convolution kernels to resample the input frames for intermediate frame interpolation~\cite{niklaus2017videoa,cheng2020video,lee2020adacof,niklaus2017videob,niklaus2021revisiting,shi2021video}. But most of these methods only consider resampling of local neighborhood patches, leading to blurry results. 

\subsection{Spatial-temporal Video Super-resolution}
ST-VSR is to increase the spatial and temporal resolution of low-resolution (LR) and low-frame-rate (LFR) videos~\cite{shahar2011space}. The key challenge lies in sufficiently exploiting the spatial-temporal information of input frames. For instance, Shechtman \textit{et al.}~\cite{shechtman2005space} adopt a directional spatial-temporal smoothness regularization to constrain high spatial-temporal resolution. However, this constrain makes it difficult model spatial-temporal correlations on complex or diverse visual patterns. Lately, learning-based methods~\cite{dutta2021efficient} attempt to decompose ST-VSR into two sub-tasks that are achieved on image space sequentially: spatial video super-resolution (S-VSR) and temporal video super-resolution (T-VSR). However, they fail to utilize intrinsic relations between S-VSR and T-VSR. Recently, some studies~\cite{xiang2021zooming,xu2021temporal} show one-stage based ST-VSR methods are significantly better than two-stage based ST-VSR methods on effectiveness and efficiency. STARnet~\cite{haris2020space}, MBnet~\cite{zhou2021video} and CycMu-Net~\cite{hu2022spatial} propose a mutual learning strategy for ST-VSR, which makes full use of spatial-temporal information by jointly learning S-VSR and T-VSR via iterations. %However, they are still require more parameters and time-consuming, especially high resolution. 
Zooming Slow-Mo~\cite{xiang2020zooming} propose to firstly capture local temporal contexts for intermediate feature interpolation by deformable convolution~\cite{zhu2019deformable}, then explore global temporal contexts to further build temporal correlations by bidirectional deformable ConvLSTM~\cite{shi2015convolutional}, and finally reconstructs high-resolution (HR) and high-frame-rate (HFR) spatial-temporal videos by a reconstruction network. Inspired by~\cite{xiang2020zooming}, Xu \textit{et al.}~\cite{xu2021temporal} further propose a locally temporal feature comparison module to extract local motion cues for refinement, achieving better performances on two public datasets. However, these one-stage based methods are procedure redundant due to ignorance of effective combination of common pipelines.

\section{Proposed Method}
\subsection{Framework Overview}
Spatial-temporal video super-resolution (ST-VSR) aims to reconstruct high-resolution (HR) and high-frame-rate (HFR) video sequences $[I_{t}^H]_{t=1}^{2n+1}$ from given low-resolution (LR) and low-frame-rate (LFR) inputs $[I_{2t-1}^L]_{t=1}^{n+1}$. Figure~\ref{framework} shows the pipeline of our proposed YOGO (You Only aliGn Once) algorithm, which mainly consists of four seamless components: Feature Extraction, Alignment with Bidirectional Interaction Propagation Module (BIPM), Fusion, Reconstruction.
%bidirectional interaction propagation (BIP) involves a one-stage and end-to-end architecture by sharing the common pipelines in T-VSR and S-VSR, which mainly consists of four seamless components: \textbf{Feature Extraction}, \textbf{Alignment with Bidirectional Interaction Propagation (BIP)}, \textbf{Fusion}, \textbf{Reconstruction}.

Given LR and LFR video sequences $[I_{2t-1}^L]_{t=1}^{n+1}$, a feature extraction (FE) module, involving one convolution layer and five residual blocks~\cite{he2016deep}, is used to project input frames into feature space to generate the initial representation ($[F_{t}^{L}]_{t=1}^{2n+1}$). Followed by a Bidirectional Interaction Propagation Module (BIPM), the temporal information between these frames is sufficiently exploited to estimate the motion. In particular, BIPM allows the network to aggregate the feature representation from the past, current and future frames. More specifically, recurrent cell with backward inference (RC-BI) learns the temporal relations from future to past frame, and  packages them into a hidden unit while recurrent cell with forward inference (RC-FI) can exploit the spatial-temporal information  from all frames by implicitly aligning and updating the packaged hidden unit. To further aggregate the spatial and temporal information, a hybrid fusion module (HFM) containing multiple hybrid fusion blocks (HFBs) is designed to combine the outcomes of these two inferences, and two pixel-shuffle layers~\cite{shi2016real} are used to yield the final HR and HFR solution $[I_{t}^H]_{t=1}^{2n+1}$.

%\textbf{Alignment with Bidirectional Interaction Propagation (BIP):} To make full use of temporal information, bidirectional interaction propagation module (BIPM) is designed to firstly gather hidden states from future frames to past frames via backward propagation, and then combine hidden states from backward propagation to collect temporal information from all frames via forward propagation, followed by implicitly aligning corresponding hidden states via recurrent cell with backward interactive propagation (RC-BIP) and recurrent cell with forward interactive propagation (RC-FIP), respectively.
%\textbf{Fusion:} We propose hybrid fusion block (HFB) to assign backward and forward inference to learn structures and details from temporal information, and further progressively aggregate and distill structures and details components for spatial information refinement. \textbf{Reconstruction:} We produce the final HR and HFR frames $[I_{t}^H]_{t=1}^{2n+1}$, structures $[S_{t}^H]_{t=1}^{2n+1}$ and details $[D_{t}^H]_{t=1}^{2n+1}$  via two pixel-shuffle layers~\cite{shi2016real} by supervising them. 

\tabcolsep=0.5pt
\begin{figure}[t]
     \centering
     \includegraphics[width=1.0\linewidth]{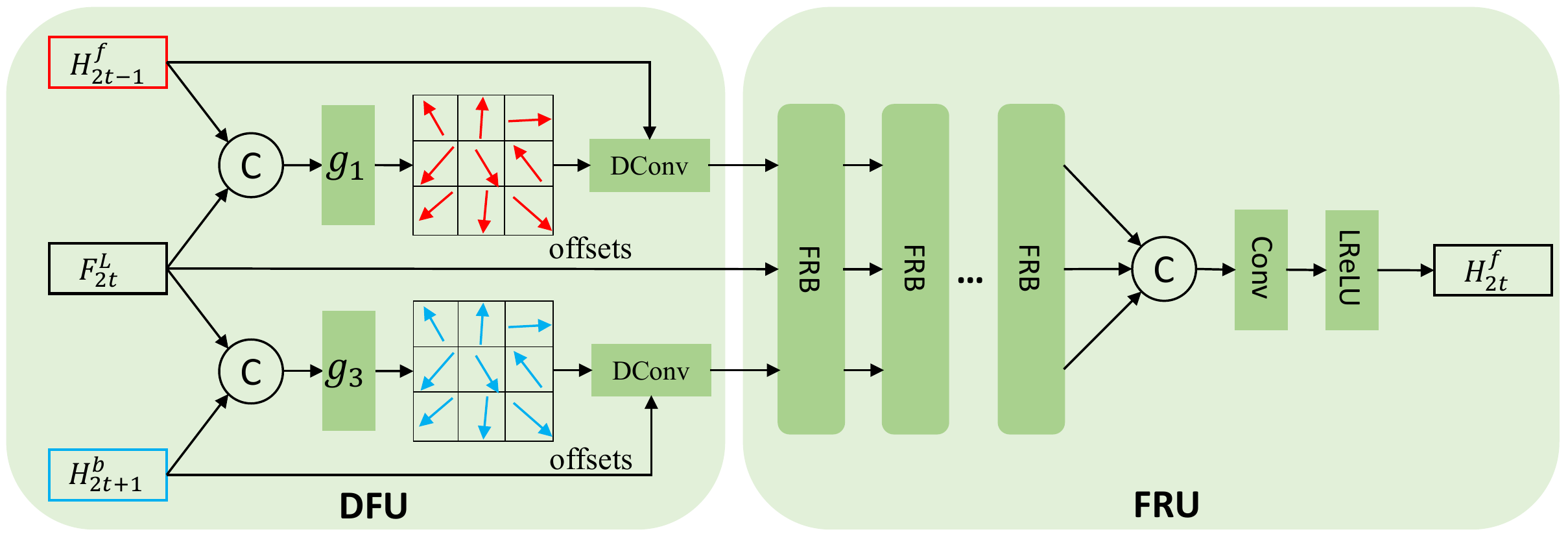}
     \caption{Architecture of the proposed recurrent cell with forward inference (RC-FI). Note that the backward inference (RC-BI) shares the similar framework to RC-FI, but takes $H_{2t+1}^b$ and $F_{2t}^L$ as inputs.}
     \label{framework1}
     %\vspace{-6mm}
\end{figure}

\subsection{Bidirectional Interactive Propagation Module}
%Since advanced spatial-temporal video super-resolution (ST-VSR) methods sequentially perform spatial video super-resolution (S-VSR) and temporal video super-resolution (T-VSR) on image or feature spaces and vice versa. We find that they fail to fully learn the intrinsic correlations of bidirectional motions for spatial reconstruction, and alignment and fusion pipelines are repetitive in T-VSR and S-VSR. %(\textit{e.g.} some methods~\cite{xiang2020zooming,xu2021temporal} conduct motion estimation $O_{{2t-1}{\rightarrow}{2t}}$ between $I_{2t-1}$ and $I_{2t+1}$ to align intermediate representation, then further perform motion estimation $O_{{2t-1}{\rightarrow}{2t}}$ between $I_{2t-1}$ and $I_{2t}$ to align intermediate representations for high-resolution (HR) reconstruction). 
Advanced spatial-temporal video super-resolution (ST-VSR) methods sequentially perform spatial video super-resolution (S-VSR) and temporal video super-resolution (T-VSR) on image or feature spaces. However, individual alignment and fusion operations are adopted in these methods to fuse spatial and temporal information. Besides the structure redundancy, motion estimation and alignment in T-VSR and S-VSR are independent and lack interactions, which hinders accurate spatial-temporal correlation learning.
To alleviate this issue, we propose a novel bidirectional interactive propagation module (BIPM), where the backward and forward inferences share the alignment pipeline to implicitly learn the spatial-temporal correlations from all frames. More specifically, a recurrent cell with backward inference (RC-BI) learns the temporal relations from future to past frame, and packages them into a hidden unit while a recurrent cell with forward inference (RC-FI) can exploit the spatial-temporal information from all frames by implicitly aligning and updating the packaged hidden unit. For convenience, RC-BI and RC-FI have the similar framework, composed of a dynamic filter unit (DFU) and a fusion residual unit (FRU). The proposed DFU explores short-term temporal correlations by aligning hidden states, while FRU further explores long-term variations of the whole video by utilizing the current input representation and aligned hidden states. This two-stage temporal feature alignment scheme accurately and sufficiently aggregates the spatial-temporal features.     
%which consists of two recurrent cells with bidirectional inferences scheme to share alignment pipeline between S-VSR and T-VSR. Recurrent cell with backward inference (RC-BI)learns the temporal relations from future to past frame, and packages them into a hidden unit while recurrent cell with forward inference (RC-FI) can exploit the spatial-temporal information from all frames by implicitly aligning and updating the packaged hidden unit. Specifically, the proposed recurrent cell composed of two subnetworks: dynamic filter network (DFNet) and fusion residual network (FRNet). 

\tabcolsep=0.5pt
\begin{figure}[t]
     \centering
     \includegraphics[width=1.0\linewidth]{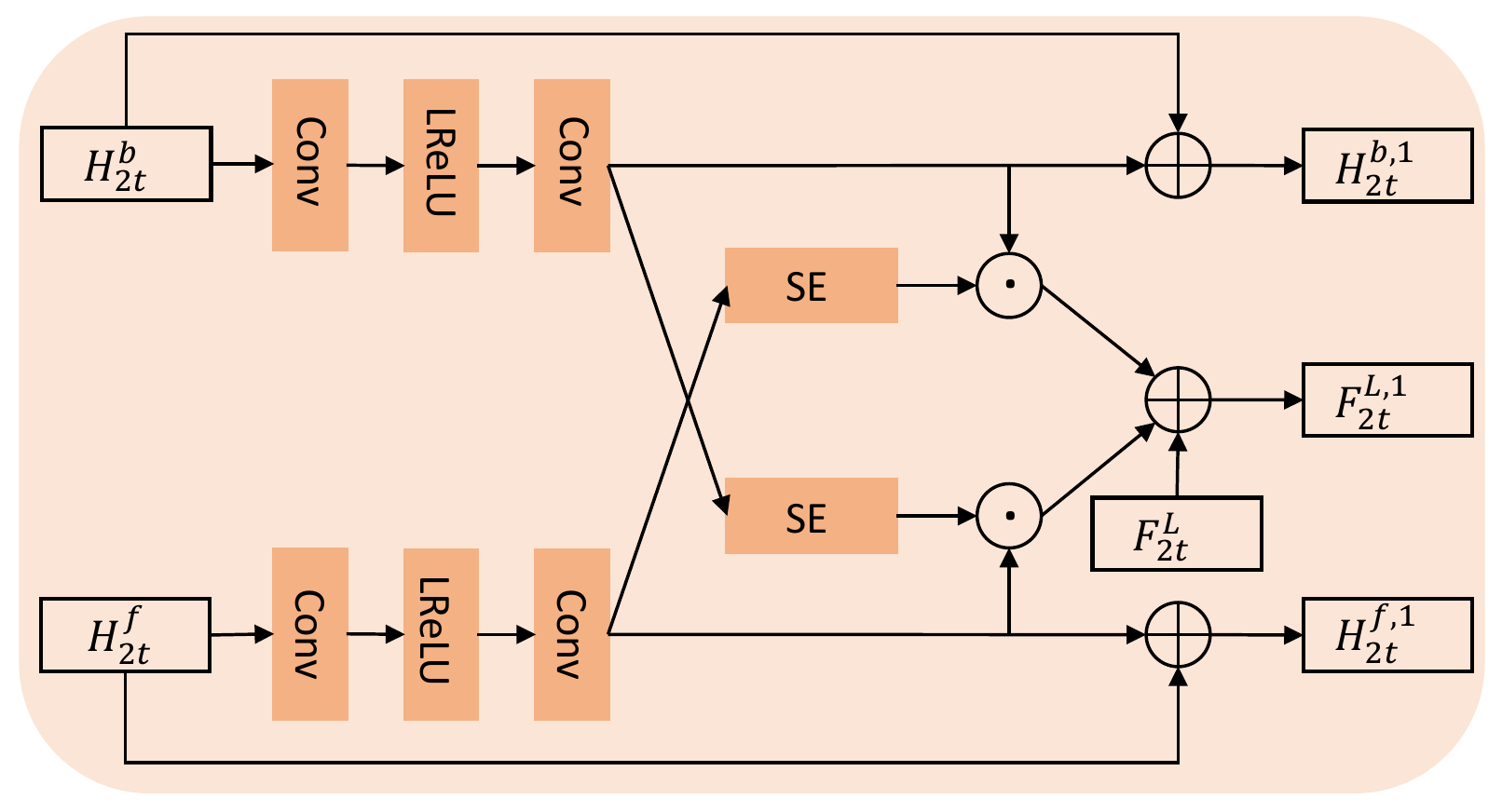}
     %\vspace{-2mm}
     \caption{Architecture of the proposed hybrid fusion block (HFB). SE denotes squeeze-and-excitation networks~\cite{hu2018squeeze}.}
     %\vspace{-5mm}
     \label{framework2}
\end{figure}

\textbf{DFU:} 
Considering the feature redundancy and alignment errors between the different frames and backward-forward inference, we introduce a dynamic filter unit (DFU) to first learn offsets among the current input and hidden states. It provides the guidance to distill and fuse the informative components while mitigating the accumulated errors.
%To make full use of temporal information in hidden state, we propose a dynamic filter network (DFNet) to estimate offsets between present input representations and hidden states, which allows the present representations to sample useful information and suppress outdated information from hidden states for mitigation of accumulated errors.
%Inspired in~\cite{jia2016dynamic,niklaus2017videob,xiang2020zooming}, as shown in Figure~\ref{framework1}, 
Taking RC-FI as an example, as shown in Figure~\ref{framework1}, the proposed DFU takes past hidden state $H_{2t-1}^f$ and input representation $F_{2t}^L$ , as well as the future hidden state $H_{2t+1}^b$ from backward inference as input, and predicts the offsets via the offset estimator, which guide the deformable convolutions to capture the most related components. The implicit alignment in DFU can be expressed as
%by employing the pyramid, cascading and deformable module ~\cite{wang2019edvr} to estimate offsets $\Delta{P_{1}}$ and $\Delta{P_{2}}$, which can dynamically sample multiple pixels of interest and linearly combines the sampled values via deformable convolution for accurate and implicit alignment. The process can be defined as follow:
\begin{align}
\begin{split}
\Delta{P}_{1} &= {g_1}([H_{2t-1}^f,F_{2t}^L]),\\
\Delta{P}_{2} &= {g_3}([H_{2t+1}^b,F_{2t}^L]),\\
\hat{H}_{2t-1}^f &= DConv(H_{2t-1}^f,\Delta{P}_{1}),\\
\hat{H}_{2t+1}^b &= DConv(H_{2t+1}^b,\Delta{P}_{2}),\\
\end{split}
\end{align}
where $\Delta{P}_{1}$ and $\Delta{P}_{2}$ are the learnable offsets via the offset estimator ($g_1(\cdot)$ and $g_3(\cdot)$)~\cite{wang2019edvr}. $DConv(\cdot)$ denotes the deformable convolution. %$[,]$ denotes the channel-wise concatenation. 
%$g_1(\cdot)$ and $g_3(\cdot)$ are offsets estimator~\cite{wang2019edvr} with shared parameters.

\tabcolsep=0.5pt
\begin{table*}[t]
\setlength\tabcolsep{4pt}
\centering
\caption{Quantitative comparisons of our results and two-stage ST-VSR methods on Vid4 and Vimeo90K datasets. The best and second best results are highlighted in \textcolor{red}{red} and \textcolor{blue}{blue}, respectively. "Ours (56)" denotes YOGO with the filter number as 56.}
%\vspace{-2mm}
\resizebox{1.0\textwidth}{!}
{
\smallskip\begin{tabular}{|cc|cc|cc|cc|cc|c|}
\hline
\multicolumn{1}{|c}{VFI}&\multicolumn{1}{c}{VSR}& \multicolumn{2}{|c}{Vid4} & \multicolumn{2}{|c}{Vimeo90K-Fast}& \multicolumn{2}{|c}{Vimeo90K-Medium} & \multicolumn{2}{|c}{Vimeo90K-Slow}& \multicolumn{1}{|c|}{Parameters}\cr
Method&Method&PSNR$\uparrow$&SSIM$\uparrow$&PSNR$\uparrow$&SSIM$\uparrow$&PSNR$\uparrow$&SSIM$\uparrow$&PSNR$\uparrow$&SSIM$\uparrow$&(millions)$\downarrow$ \cr \hline
SuperSloMo&Bicubic&22.84&0.5772&31.88&0.8793&29.94&0.8477&28.37&0.8102&\textcolor{blue}{19.8}\cr
SuperSloMo&RCAN&23.80&0.6397&34.52&0.9076&32.50&0.8884&30.69&0.8624&19.8+16.0\cr
SuperSloMo&RBPN&23.76&0.6362&34.73&0.9108&32.79&0.8930&30.48&0.8584&19.8+12.7\cr
SuperSloMo&EDVR&24.40&0.6706&35.05&0.9136&33.85&0.8967&30.99&0.8673&19.8+20.7
\cr
\hline
SepConv&Bicubic&23.51&0.6273&32.27&0.8890&30.61&0.8633&29.04&0.8290&21.7\cr
SepConv&RCAN&24.92&0.7236&34.97&0.9195&33.59&0.9125&32.13&0.8967&21.7+16.0\cr
SepConv&RBPN&26.08&0.7751&35.07&0.9238&34.09&0.9229&32.77&0.9090&21.7+12.7\cr
SepConv&EDVR&25.93& 0.7792& 35.23 &0.9252& 34.22& 0.9240 &32.96& 0.9112 & 21.7+20.7
\cr
\hline
DAIN&Bicubic&23.55& 0.6268 &32.41 &0.8910 &30.67 &0.8636 &29.06 &0.8289 &24.0\cr
DAIN&RCAN&25.03& 0.7261 &35.27 &0.9242 &33.82 &0.9146 &32.26 &0.8974 &24.0+16.0\cr
DAIN&RBPN&25.96 &0.7784 &35.55 &0.9300 &34.45 &0.9262 &32.92 &0.9097 &24.0+12.7\cr
DAIN&EDVR&\textcolor{blue}{26.12} &\textcolor{blue}{0.7836} & \textcolor{blue}{35.81} &\textcolor{blue}{0.9323} &\textcolor{blue}{34.76} &\textcolor{blue}{0.9281} & \textcolor{blue}{33.11} & \textcolor{blue}{0.9119} &24.0+20.7
\cr
\hline
\multicolumn{2}{|c|}{Ours (56)}&\textcolor{red}{26.28}&\textcolor{red}{0.7996}&\textcolor{red}{36.76}&\textcolor{red}{0.9397}&\textcolor{red}{35.32}&\textcolor{red}{0.9349}&\textcolor{red}{33.33}&\textcolor{red}{0.9134}&\textcolor{red}{9.5}\cr
\hline

\end{tabular}
}

\label{Two-stage}
\end{table*}

\tabcolsep=0.5pt
\begin{table*}[t]
\setlength\tabcolsep{4pt}
\centering

\caption{Quantitative comparisons of our results and one-stage ST-VSR methods on Vid4 and Vimeo90K datasets. "Ours (56)" and "Ours (64)" denote YOGO with the filter number as 56 and 64, respectively. The total runtime is measured on the entire Vid4 dataset~\cite{liu2011bayesian}, Note we input four LR image with the resolution of 180$\times$144 to test FLOPS.}
%\vspace{-2mm}
\resizebox{1.0\textwidth}{!}
{
\smallskip\begin{tabular}{|c|cc|cc|cc|cc|c|c|c|}
\hline
\multicolumn{1}{|c}{ST-VSR}& \multicolumn{2}{|c}{Vid4} & \multicolumn{2}{|c}{Vimeo90K-Fast}& \multicolumn{2}{|c}{Vimeo90K-Medium} & \multicolumn{2}{|c}{Vimeo90K-Slow}& \multicolumn{1}{|c}{Speed}&\multicolumn{1}{|c}{FLOPs} & \multicolumn{1}{|c|}{Parameters}\cr
Method&PSNR$\uparrow$&SSIM$\uparrow$&PSNR$\uparrow$&SSIM$\uparrow$&PSNR$\uparrow$&SSIM$\uparrow$&PSNR$\uparrow$&SSIM$\uparrow$&(fps)$\uparrow$&(T)$\downarrow$&(millions)$\downarrow$ \cr \hline
\multicolumn{1}{|c|}{STARnet}&26.06 &\textcolor{red}{0.8046} &36.19 &0.9368 &34.86 &0.9356 &33.10 &\textcolor{red}{0.9164} &14.92&27.926&111.6
\cr
\multicolumn{1}{|c|}{Zooming Slow-Mo}&{26.31} &0.7976 &{36.81} &{0.9415} &{35.41} &{0.9361} &{33.36} &{0.9138} &\textcolor{red}{17.34}&1.766 &\textcolor{blue}{11.1}\cr
\multicolumn{1}{|c|}{TMNet}&\textcolor{red}{26.43} &{0.8016} &\textcolor{red}{37.04} &\textcolor{red}{0.9435} &\textcolor{red}{35.60} &\textcolor{red}{0.9380} &\textcolor{red}{33.51} &\textcolor{blue}{0.9159} &{15.62}&1.874 &{12.3}\cr
\multicolumn{1}{|c|}{Ours (56)}&{26.28}&{0.7996}&{36.76}&{0.9397}&{35.32}&{0.9349}&{33.33}&{0.9134}&\textcolor{blue}{16.72}&\textcolor{red}{1.148}&\textcolor{red}{9.5}\cr
\multicolumn{1}{|c|}{Ours (64)}&\textcolor{blue}{26.34}&\textcolor{blue}{0.8022}&\textcolor{blue}{36.93}&\textcolor{blue}{0.9416}&\textcolor{blue}{35.55}&\textcolor{blue}{0.9365}&\textcolor{blue}{33.44}&{0.9150}&{15.87}&\textcolor{blue}{1.470}&{12.1}\cr
%\multicolumn{1}{|c|}{Ours (80)}&{0}&{0}&{0}&{0}&{0}&{0}&{0}&{0}&{0}&{0}\cr
\hline
\end{tabular}
}
%\vspace{-4mm}
\label{One-stage}
\end{table*}

\tabcolsep=0.5pt
\begin{figure*}[!htb]
	\centering
\footnotesize{
		\begin{tabular}{ccccccc}
			\includegraphics[width=0.17\textwidth]{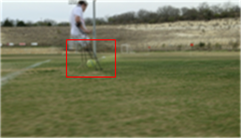} &
			\includegraphics[width=0.17\textwidth]{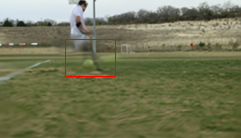} &
			\includegraphics[width=0.17\textwidth]{./figure5/seprrpn.png} &
			\includegraphics[width=0.17\textwidth]{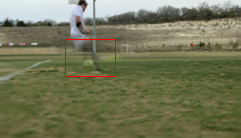} &
			\includegraphics[width=0.17\textwidth]{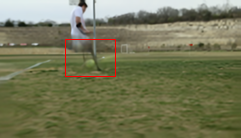} \\
			\includegraphics[width=0.17\textwidth]{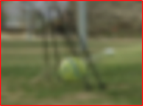} &
			\includegraphics[width=0.17\textwidth]{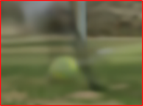} &
			\includegraphics[width=0.17\textwidth]{./figure5/seprrpn1.png} &
			\includegraphics[width=0.17\textwidth]{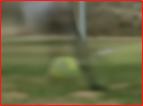} &
			\includegraphics[width=0.17\textwidth]{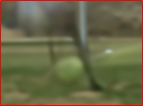} \\
			\textbf{Overlayed LR} & \textbf{DAIN+Bicubic} &\textbf{SepConv+RBPN} & \textbf{SepConv+EDVR} &\textbf{DAIN+EDVR}\\
		    & (24.92/0.8441)& (28.62/0.9173) & (28.67/0.9187) & ({28.86}/{0.9201}) \\
		    \includegraphics[width=0.17\textwidth]{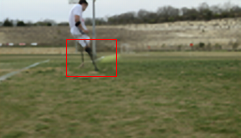} &
			\includegraphics[width=0.17\textwidth]{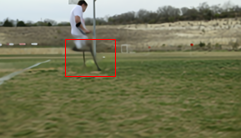} &
			\includegraphics[width=0.17\textwidth]{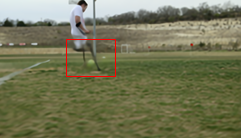} &
			\includegraphics[width=0.17\textwidth]{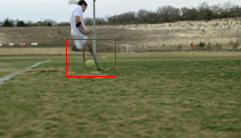} &
			\includegraphics[width=0.17\textwidth]{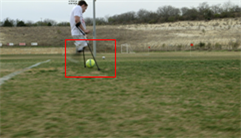} \\
			
			\includegraphics[width=0.17\textwidth]{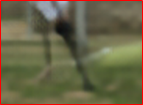} &
			\includegraphics[width=0.17\textwidth]{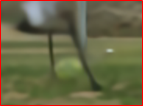} &
			\includegraphics[width=0.17\textwidth]{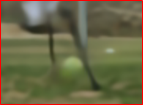} &
			\includegraphics[width=0.17\textwidth]{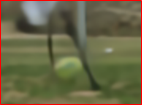} &
			\includegraphics[width=0.17\textwidth]{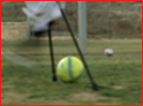} \\
			\textbf{STARnet} & \textbf{Zooming Slow-Mo} &\textbf{TMNet} & \textbf{Ours} &\textbf{Ground-Truth}\\
		    (25.43/0.8586) &(29.57/0.9283)& (\textcolor{blue}{29.63}/ \textcolor{blue}{0.9293}) & (\textcolor{red}{29.74}/\textcolor{red}{0.9301})&  \\
	\end{tabular}}
	%\vspace{-4mm}
   \caption{\small Visual comparisons with state-of-the-art two-stage and one-stage based methods on \textbf{Vimeo90K} dataset.
   }
\label{Two-stage visual}
\end{figure*}

\tabcolsep=0.5pt
\begin{figure*}[!htb]
	\centering
\footnotesize{
		\begin{tabular}{ccccccc}
			\includegraphics[width=0.13\textwidth]{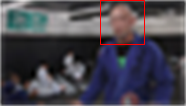} &
			\includegraphics[width=0.13\textwidth]{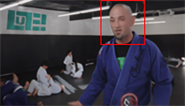} &
			\includegraphics[width=0.13\textwidth]{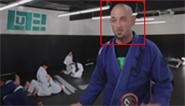} &
			\includegraphics[width=0.13\textwidth]{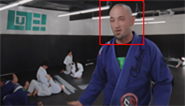} &
			\includegraphics[width=0.13\textwidth]{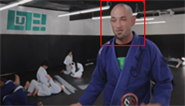} &
			\includegraphics[width=0.13\textwidth]{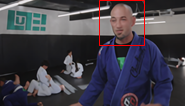} &
			\includegraphics[width=0.13\textwidth]{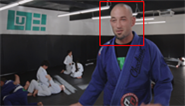} \\
			
			\includegraphics[width=0.13\textwidth]{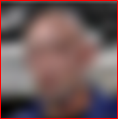} &
			\includegraphics[width=0.13\textwidth]{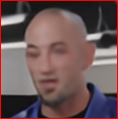} &
			\includegraphics[width=0.13\textwidth]{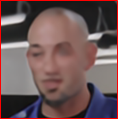} &
			\includegraphics[width=0.13\textwidth]{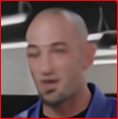} &
			\includegraphics[width=0.13\textwidth]{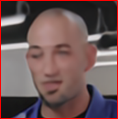} &
			\includegraphics[width=0.13\textwidth]{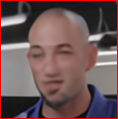} &
			\includegraphics[width=0.13\textwidth]{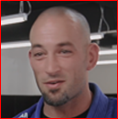} \\
			\textbf{Overlayed LR} & \textbf{Model (a)} & \textbf{Model (b)} & \textbf{Model (c)} & \textbf{Model (d)} & \textbf{Model (e)}& \textbf{Ground-Truth} \\
			& \textbf{(30.15/0.9328)} & \textbf{(31.26/0.9476)} & \textbf{(32.51/0.9593)} & \textbf{(\textcolor{blue}{33.01}/\textcolor{blue}{0.9615})} & \textbf{(\textcolor{red}{33.20}/\textcolor{red}{0.9634})} \\
	\end{tabular}}
	%\vspace{-4mm}
   \caption{\small Visual comparisons of five variants for the ablation studies \textbf{Vimeo90K} dataset.
   }
\label{ablation visual}
\end{figure*}

\tabcolsep=0.5pt
\begin{figure*}[!htb]
	\centering
\footnotesize{
		\begin{tabular}{ccccccc}
			\includegraphics[width=0.13\textwidth]{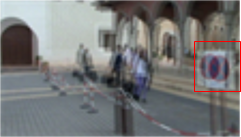} &
			\includegraphics[width=0.13\textwidth]{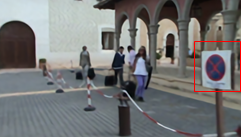} &
			\includegraphics[width=0.13\textwidth]{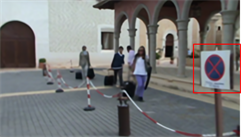} &
			\includegraphics[width=0.13\textwidth]{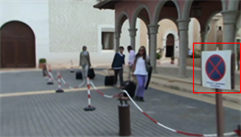} &
			\includegraphics[width=0.13\textwidth]{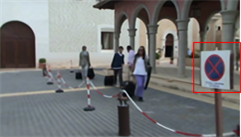} &
			\includegraphics[width=0.13\textwidth]{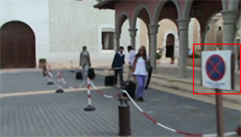} &
			\includegraphics[width=0.13\textwidth]{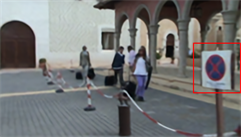} \\
			\includegraphics[width=0.13\textwidth]{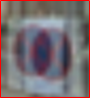} &
			\includegraphics[width=0.13\textwidth]{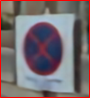} &
			\includegraphics[width=0.13\textwidth]{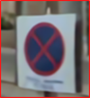} &
			\includegraphics[width=0.13\textwidth]{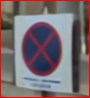} &
			\includegraphics[width=0.13\textwidth]{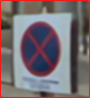} &
			\includegraphics[width=0.13\textwidth]{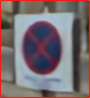} &
			\includegraphics[width=0.13\textwidth]{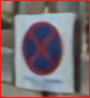} \\
			\textbf{Overlayed LR} & \textbf{0+10} & \textbf{2+8} &\textbf{4+6} & \textbf{6+4}& \textbf{8+2}& \textbf{10+0} \\
			& (28.70/0.8651) & (\textcolor{blue}{29.77}/\textcolor{blue}{0.8864}) &(\textcolor{red}{29.81}/\textcolor{red}{0.8858}) &(29.69/0.8839) &   (28.69/0.8621)&(28.59/0.8610)  \\
	\end{tabular}}
	%\vspace{-4mm}
   \caption{\small Visual comparisons by djusting FRB in FRU from RC-BI and RC-FI on \textbf{Vimeo90K} dataset.
   }
\label{frb visual}
\end{figure*}

\tabcolsep=0.5pt
\begin{figure}[t]
	\centering
\footnotesize{
		\begin{tabular}{ccc}
			\includegraphics[width=0.33\linewidth]{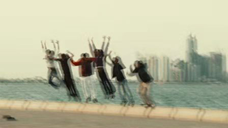} &
			\includegraphics[width=0.33\linewidth]{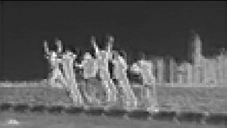} &
			\includegraphics[width=0.33\linewidth]{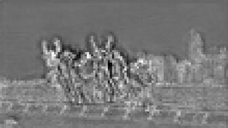} \\
			\includegraphics[width=0.33\linewidth]{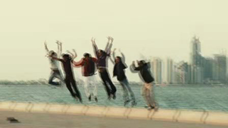} &
			\includegraphics[width=0.33\linewidth]{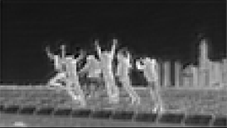} &
			\includegraphics[width=0.33\linewidth]{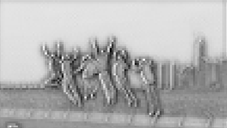} \\
			\textbf{Overlayed LR} & \textbf{DFNet} & \textbf{FRNet}\\
	\end{tabular}}
%\vspace{-3mm}
   \caption{\small Feature maps of DFU and FRU outputs from RC-FI have been visualized using same grayscale colormap. The first rows indicates overlayed the second and third LR frames, aligned feature of DFU from the second frame, and aligned feature of FRU from the second frame, respectively. The second rows indicates overlayed the third and fourth LR frames, aligned feature of DFU from the fourth frame, and aligned feature of FRU from the fourth frame, respectively.}
   %\vspace{-6mm}
\label{specific visual}
\end{figure}

\textbf{FRU:} 
DFU focuses on the short-term temporal correlations between the adjacent frames, rendering itself unfit for the occluded scenes or large motion, which may benefit from the long-term correlation learning. To this end, a fusion residual unit (FRU) is introduced to aggregate the spatial-temporal information from all frames. Specifically, FRU first takes aligned hidden states ($\hat{H}_{2t-1}^f$ and $\hat{H}_{2t+1}^b$) and current representation ($F_{2t}^L$) as inputs, where a series of fusion residual blocks (FRB) are introduced to learn the individual representation of current frame while eliminating the spatial-temporal redundancy through $1\times1$ channel fusion~\cite{yi2019progressive,yi2020progressive}. The outcomes of all branches can be combined to yield a final solution ($H_{2t}^f$) to the predicted frame.

%DFU only considers short-term temporal correlations by aligning present representations. However, when the inter-frame motions are large or complex scenes are encountered (\textit{e.g.} occlusion), DFU lacks the capability of modeling motions over the whole video. To overcome this issue by further exploring long-term temporal temporal correlations, the proposed FRU firstly generates residual information to fuse and distill temporal information from hidden states $\hat{H}_{2t-1}^f$ and $\hat{H}_{2t+1}^b$ and present representations $F_{2t}^L$  by a series of fusion residual blocks (FRB), then these three representations of the channel concatenation are fused to generate the final output $H_{2t}^f$ by a convolution layer. 

%In addition, For RC-BI, we input $H_{2t+1}^b$ and $F_{2t}^L$into RC-BI to obtain $H_{2t}^b$, since RC-FI contains more temporal information than RC-BI, we naturally restrict RC-BI to learn low-frequency structures and RC-FI to study high-frequency details by supervising them (The specific details are in Section 3.3).

%For the backward inference, only $H_{2t+1}^b$ and $F_{2t}^L$ are packed into RC-BI to learn the embedding representation of predicted frame.  
To simplify the learning procedure, inspired by the divide-and-conquer strategy, the backward and forward inferences are assigned to learn specific tasks, where RC-BI focuses on the global structure while RC-FI aims to refine the textures. It has been verified that refining the details with past, current and future representation in RC-FI can gain better performance than that of RC-BI. (More details are included in Section 3.3.)

\subsection{Hybrid Fusion Module}
%To fully utilize structures and details from temporal information for recovering different difficulties in spatial high- and low-frequency information. Inspired by~\cite{isobe2020video} which promotes information exchange between structures and details components for better spatial reconstruction. However, this fusion does not consider the preservation of modality-specific properties (\textit{i.e.} structures and details), leading to two modes interfering with each other. Thus, we propose multiple hybrid fusion blocks (HFBs) to progressively recover spatial high- and low-frequency information by exploring shared information between structures and details components. while keeping modality-specific properties to assign backward and forward inferences for structures and details learning by supervising them.

To sufficiently aggregate the spatial-temporal representation, the outputs ($H_{2t}^b$ and $H_{2t}^f$) of backward and forward inferences as well as the initial features ($F_{2t}^L$) are packed into a hybrid fusion module (HFM), where the individual components are progressively fused via multiple hybrid fusion blocks (HFBs) to eliminate the redundancy and yield the final prediction. Taking the first HFB as an example in Figure~\ref{framework2},  HFB takes the outputs of RC-BI and RC-FI as inputs, and applies two branches to deeply characterize the specific features, expressed as  
%$S_{2t}^{L_{m}}$ and $D_{2t}^{L_{m}}$ (Here $m$ is initialized to 0. In fact, $m=0,1...,M$ denotes the number of HFB). The proposed HFB has 3 branches, where the upper and lower branches maintain details- and structures-specific properties by learning residuals, which can be described as follow:
\begin{align}
\begin{split}
% &{S}_{2t}^{L_{m+1}} = {S}_{2t}^{L_{m}} + {R_1}({S}_{2t}^{L_{m}}),\\
% &{D}_{2t}^{L_{m+1}} = {D}_{2t}^{L_{m}} + {R_2}({D}_{2t}^{L_{m}}),\\
&H_{2t}^{b,1} = H_{2t}^b + R_1(H_{2t}^b),\\
&H_{2t}^{f,1} = H_{2t}^f + R_2(H_{2t}^f),\\
\end{split}
\end{align}
where $R_1(\cdot)$ and $R_2(\cdot)$ denote residual blocks~\cite{he2016deep}.

The middle branch is to exploit the correlations between the backward and forward inferences via cross-attention, the feature map from one modality can be used to enhance another modality. In this way, the spatial-temporal information from all frames is sufficiently aggregated for a better spatial reconstruction. Thus, this process is described as 
\begin{align}
\begin{split}
F_{2t}^{L,1} = F_{2t}^L &+ SE_1(R_1(H_{2t}^b))\odot R_2(H_{2t}^f)\\ &+ SE_2(R_2(H_{2t}^f)) \odot R_1(H_{2t}^b),\\
\end{split}
\end{align}
where $SE_1(\cdot)$ and $SE_2(\cdot)$ denote squeeze-and-excitation networks~\cite{hu2018squeeze}.

\subsection{Reconstruction and Optimization}
Finally, a reconstruction module (Recon) is designed to output the HR (4$\times$ ) and HFR (2$\times$ ) video $[I_t^H]_{t=1}^{2n+1}$, involving two pixel-shuffle layers~\cite{shi2016real} and a sequence of ``Conv-LeakyReLU-Conv'' operations. Meanwhile, specific representations of backward and forward inferences are projected into the image space to generate the corresponding structures ($[S_t^H]_{t=1}^{2n+1}$) and details ($[D_t^H]_{t=1}^{2n+1}$). Then, the loss functions on the predicted frames, structure and detail images are expressed as 
\begin{align}
\begin{split}
L_r = \sum_{t=1}^{2n+1}(\rho(I_t^H-I_t^{GT}) + \rho(D_t^H-D_t^{GT}) + \rho(S_t^H-S_t^{GT})) ,
\end{split}
\end{align}
where $I_t^{GT}$, $D_t^{GT}$, $S_t^{GT}$ refer to the corresponding ground-truth video frames, details and structures components, where the detail components denote the residue between the bicubic sampling (the structural components) and the ground-truth video frames $I_t^{GT}$.
%downsampling and upsampling ground-truth video frames $I_t^{GT}$. Ground-truth detail components $D_t^{GT}$ can be then computed as the difference between ground-truth frame $I_t^{GT}$ and structures components $S_t^{GT}$. 
$\rho = \sqrt{(x^2 + w^2)}$ is the Charbonnier penalty function where constant $w$ is set to  $10^{-3}$~\cite{charbonnier1994two}.

\subsection{Implementation Details} In our study, the training images are randomly cropped into small patches with a fixed size of 112$\times$64 and randomly rotated and flipped. We take out the odd-indexed 4 frames as LR and LFR inputs, and the corresponding consecutive HR 7-frame for supervision. Specifically, the batch size is set to 10. We trained our proposed YOGO using Pytorch 1.9 with four NVIDIA Tesla V100 and adopt AdaMax optimizer~\cite{kingma2014adam}, where the initial learning is set to $10^{-4}$ with the decay rate of 0.1 at every 30 epochs till 70 epochs.

\section{EXperiments and Analysis}
\subsection{Datasets and Metrics}
%\textbf{Datasets.} 
Similar to~\cite{xiang2021zooming}, the \textbf{Vimeo90K trainset} is used to train our YOGO and other compared methods for fairness. This dataset consists of more than 60,000 7-frame training video sequences with the resolution of 448$\times$256~\cite{xue2019video}. In addition, \textbf{Vid4}~\cite{liu2011bayesian} and \textbf{Vimeo90K} testsets are used as the evaluation datasets. Following~\cite{xiang2020zooming}, \textbf{Vimeo90K} testsets are split into three subsets of fast, medium and slow motion, including 1225, 4977 and 1613 video sequences, respectively. In this study, before passed into the network, the odd-indexed LR frames are sampled via bicubic to generate the degraded inputs. %to predict the corresponding HR (4$\times$) and HFR (2$\times$) videos.
Two commonly used evaluation metrics, such as Peak
Signal to Noise Ratio (PSNR) and Structural Similarity (SSIM)~\cite{wang2004image}, as well as the Frame Per Second (fps) are employed for comparison. Higher fps indicates faster speed.
%\textbf{Metric.} we use Peak Signal-to-Noise Ratio (PSNR), Structural Similarity Index (SSIM)~\cite{wang2004image} and Frame Per Second (fps) for performance evaluations. Higher PSNR and SSIM indicate better performance. Higher fps indicates faster speed.

\subsection{Comparison with State-of-the-Art Methods}

We compare our proposed network with state-of-the-art two-stage based ST-VSR methods. We perform T-VSR by SuperSloMo~\cite{jiang2018super}, SepConv~\cite{niklaus2017videob} and DAIN~\cite{bao2019depth}, and perform S-VSR by Bicubic Interpolation, RCAN~\cite{wang2018learning},  RBPN~\cite{haris2019recurrent},
and EDVR~\cite{wang2019edvr}. In addition,  we also compare our proposed network with state-of-the-art one-stage based ST-VSR methods, including STARnet~\cite{haris2020space}, Zooming SlowMo~\cite{xiang2020zooming} and TMNet~\cite{xu2021temporal}. we test these methods on the all testsets based on public codes provided by authors. 

\textbf{Quantitative results.} Quantitative results are tabulated in Table~\ref{Two-stage} and Table~\ref{One-stage}. It is obvious that two-stage based methods require more parameters while one-stage based methods achieve better performances with only half or even fewer parameters. This is mainly attributed to the fact that two-stage based methods contain multiple redundant pipelines, like the individual feature extraction and reconstruction in T-VSR and T-VSR. %in particular, two large reconstruction network. 
Furthermore, the best two-stage methods is \textcolor{blue}{0.95dB} lower than our method YOGO (56) on Vimeo90k-Fast dataset. In addition, While the
accuracy of YOGO (64) is marginally worse than TMNet~\cite{xu2021temporal}, it is faster
by more than \textcolor{blue}{0.25fps} with only about \textcolor{blue}{78\%} of calculation cost. Compared to Zooming Slow-Mo~\cite{xiang2020zooming}, YOGO (64) achieves better results on all testsets with only about \textcolor{blue}{83\%} of calculation cost. Furthermore, YOGO (56) also achieve competitive accuracy with \textcolor{blue}{less parameters and calculation cost}. All these results validate the effectiveness of our proposed method for ST-VSR in efficiency.  

\textbf{Qualitative results.} 
Figure~\ref{Two-stage visual} provides the visual comparison of three mainstream one-stage and two-stage based ST-VSR baselines, while the quantitative results (PSNR and SSIM) are also compared.
%with their PSNR and SSIM values in Figure~\ref{Two-stage visual}.
It is observed that two-stage based ST-VSR methods tend to produce blurry results with more artifacts (\textcolor{blue}{see the stick in the red boxes}). The main reason is that the T-VSR and S-VSR are performed independently, where the spatial-temporal correlations between two tasks are under-explored~\cite{xiang2020zooming,kang2020deep}. Compared to two-stage based methods, one-stage based methods can generate sharper results. However, these methods cannot sufficiently learn the intrinsic correlations of bidirectional motion and ignore the difficulties of recovering structures and details information for spatial reconstruction, producing blurry results %without texture information 
(\textcolor{blue}{see the ball and stick in the red boxes}). On the contrary, our proposed method can sufficiently aggregate the spatial and temporal information from all frames via bidirectional interactive learning, which is helpful for texture reconstruction. Moreover, the divide-and-conquer learning strategy allows the network to focus more on the specific representation, which is followed by a fusion and reconstruction to generate  more nature and pleasant results. (More comparisons are included in our \textcolor{blue}{supplementary document})

\tabcolsep=0.5pt
\begin{figure}[t]
	\centering
\footnotesize{
		\begin{tabular}{cccc}
			\includegraphics[width=0.25\linewidth]{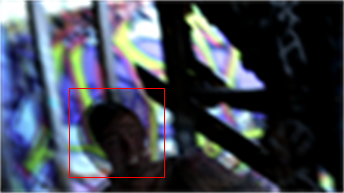} &
			\includegraphics[width=0.25\linewidth]{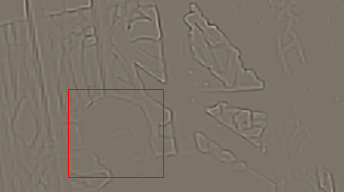} &
			\includegraphics[width=0.25\linewidth]{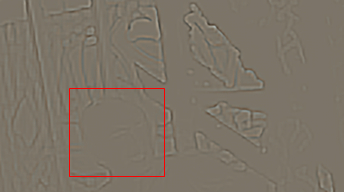} &
			\includegraphics[width=0.25\linewidth]{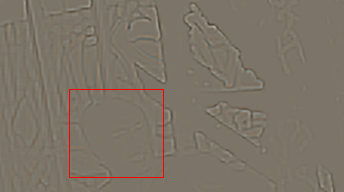} \\
			\includegraphics[width=0.25\linewidth]{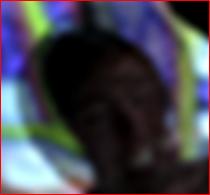} &
			\includegraphics[width=0.25\linewidth]{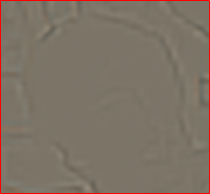} &
			\includegraphics[width=0.25\linewidth]{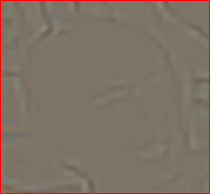} &
			\includegraphics[width=0.25\linewidth]{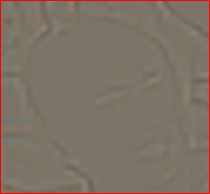} \\
			\textbf{Overlayed LR} &  \textbf{FRB (1)}  & \textbf{FRB (5)} & \textbf{FRB (9)}\\
			\includegraphics[width=0.25\linewidth]{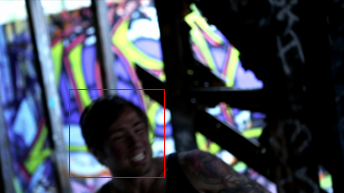} &
			\includegraphics[width=0.25\linewidth]{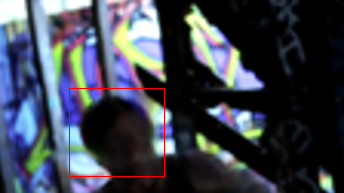} &
			\includegraphics[width=0.25\linewidth]{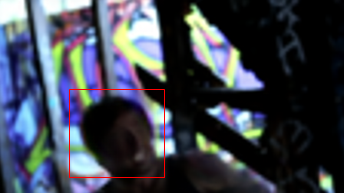} &
			\includegraphics[width=0.25\linewidth]{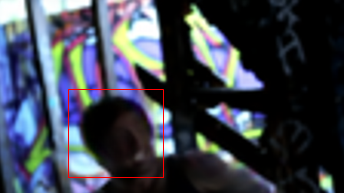} \\
			\includegraphics[width=0.25\linewidth]{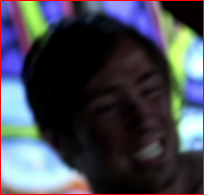} &
			\includegraphics[width=0.25\linewidth]{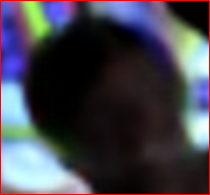} &
			\includegraphics[width=0.25\linewidth]{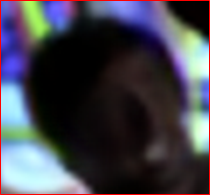} &
			\includegraphics[width=0.25\linewidth]{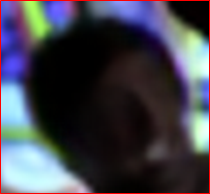} \\
			\textbf{Ground-Truth} &  \textbf{FRB (1)}  & \textbf{FRB (5)} & \textbf{FRB (9)}\\
	\end{tabular}}
	%\vspace{-4mm}
   \caption{\small Visual comparisons of details maps from different numbers of hybrid fusion blocks (HFBs) in hybrid fusion module (HFM). The third and fourth rows represent visual comparisons of structures maps from different numbers of HFBs in HFM.}
   %\vspace{-6mm}
\label{cefm visual}
\end{figure}

\subsection{Model Analysis}
In model analysis, we train all models for 50 epochs  on the \textbf{Vimeo90k} dataset with 32 $\times$ 32 cropped small patches, and test them on \textbf{Vid4} and \textbf{Vimeo90K} testsets. 

\textbf{Ablation study.} To further verify the effectiveness of different modules in our proposed network, we conduct a comprehensive ablation study on different variants.

\heading{Model (a):} We utilize a recurrent cell with only backward inference scheme to implicit align hidden states from temporal information, Then directly fuse and refine spatial information for reconstruction.

\heading{Model (b):} We utilize two recurrent cells with bidirectional inferences scheme to implicit align hidden states from temporal information, Then directly fuse and refine spatial information for reconstruction. 

\heading{Model (c):} We utilize two recurrent cells with bidirectional inferences scheme to implicit align hidden states from temporal information, Then fuse and distill structures and details to refine spatial information for reconstruction via multiple hybrid fusion blocks (HFBs).

\heading{Model (d):} We utilize two recurrent cells with bidirectional interactive inferences scheme to implicit align hidden states from temporal information, Then directly fuse and refine spatial information for reconstruction.

\heading{Model (e):} The complete version of YOGO.

The numerical comparisons and visual comparisons are shown in Table~\ref{variant ablation} and Figure~\ref{ablation visual}, we can see that \textbf{Model (b)} outperforms \textbf{Model (a)} by 0.15dB, 0.15dB and 0.14dB on \textbf{Vimeo90K-Fast}, \textbf{Vimeo90K-Medium} and \textbf{Vimeo90K-Slow} dataset, respectively, and produce clearer results (\textcolor{blue}{ the left eye of the person}), since information from past and future frame can be utilized via bidirectional inferences scheme. In addition, \textbf{Model (c)} is significantly better than \textbf{Model (b)} (\textcolor{blue}{see eyes of the person in red boxes}). This is mainly due to the fact that structure-detail fusion can focus more on the specific representation for spatial refinement. Furthermore, we observe that \textbf{Model (e)} contains more textures and further improve ST-VSR performance over \textbf{Model (c)} on three testsets, which also demonstrate that bidirectional interactive inferences scheme is more conducive to sufficiently utilize temporal information, contributing to spatial structure and detail reconstruction.

%\textbf{Effect of Bidirectional Coupled Propagation Module.} Here, we conduct detailed experiments to analyse the effectiveness of our proposed bidirectional coupled propagation module.

\heading{Different numbers of FRB in FRU from RC-BI and RC-FI:} Our proposed two recurrent cells (RCs) consists of recurrent cell with backward inference (RC-BI) and recurrent cell with forward inference (RC-FI), which individually explore structures and details from temporal information. In this section, we conduct the detailed experiments to adjust fusion residual block (FRB) in FRU from RC-BI and RC-FI to explore the optimal proportion for ST-VSR. Thus, we keep the total number of FRB fixed (10), and adjust the FRB in RC-BI (0,2,4,6,8,10) and RC-FI (10,8,6,4,2,0). As shown in Table~\ref{frb ablation}, we find neither of "0+10" and "10+0" cannot produce optimal results. This also proves that only exploring long-term temporal information from details in RC-FI or structures in RC-BI fails to mutually learn two components and cannot sufficiently and accurately utilize temporal information, affecting the recovery of spatial structures and details. %To better understand our analysis, we also visual results from different proportions of FRB in RC-BI and RC-FI in Figure~\ref{frb visual}, we can find that results from "0+10" and "10+0" have blurry structures and lack texture information (\textcolor{blue}{see words on the road sign in the red boxes}). 
On the contrary, "4+6" outperforms all other proportions in structures and details components reconstruction. This also proves that it is important to simultaneously explore the long-term structures and details from  temporal information for spatial recovery, especially details information.

\begin{table}[t]
\smallskip
\centering
\setlength\tabcolsep{4pt}

\caption{Quantitative comparisons in PSNR on the performances of different modules. Direct Fusion denotes the direct processing of temporal information in fusion process, SD Fusion denotes processing of structure and detail components from temporal information in fusion process.}
%\vspace{-2mm}
\resizebox{\columnwidth}{!}
{
\begin{tabular}{|c|c|c|c|c|c|}
\hline
Setting &\textbf{Model (a)} &\textbf{Model (b)} &\textbf{Model (c)} &\textbf{Model (d)} &\textbf{Model (e)} \\
\hline
Single Direction  &\cmark & \xmark&\xmark &\xmark &\xmark \\

Bidirection  &\xmark &\cmark &\cmark &\xmark &\xmark \\

Bidirectional Interaction  &\xmark &\xmark &\xmark&\cmark &\cmark \\
\hline
Direct Fusion  &\cmark &\cmark & \xmark &\cmark &\xmark \\

SD Fusion &\xmark &\xmark & \cmark &\xmark &\cmark \\
\hline
Vimeo90K-Fast&35.36 & 35.51& 35.71& \textcolor{blue}{36.08}&\textcolor{red}{36.18}\\
Vimeo90K-Medium&34.21 & 34.36& 34.58&\textcolor{blue}{34.86}&\textcolor{red}{34.96} \\
Vimeo90K-Slow&32.42 & 32.56& 32.74& \textcolor{blue}{32.93}&\textcolor{red}{33.02} \\
\hline

\end{tabular}
}
%\vspace{-4mm}
\label{variant ablation}
\end{table}

\heading{The Impact of DFU and FRU:} To further analyze the specific role of DFU and FRU, as shown in Figure~\ref{specific visual} and Table~\ref{BCP ablation}, we visualize bidirectional output feature maps from DFU and FRU in a gray level. We can find the former is more concerned with local motion information when exploring short-term temporal correlations, while the latter is more effective in exploring the global motion information by utilizing the long-term temporal correlations of the whole sequence, especially motion boundaries and background information. This is also in line with our expectation that DFU and FRU perform two-stage feature alignment scheme accurately aggregate the spatial-temporal information. In addition, "DFU+FRU" outperforms "FRU+DFU" on three testsets. The main reason is that performing FRU before DFU brings noisy long-term irrelevant information, confusing the short-term temporal correlations learning of DFU. This indicates that it is essential that we firstly utilize DFU and then FRU to sufficiently explore and utilize temporal information.

\begin{table}[t]
\smallskip
\centering
\setlength\tabcolsep{8pt}
\caption{: PSNR (dB) evaluated by adjusting FRB in FRU. “6+4” denotes setting 6 FRBs in RC-BI and 4 in RC-FI.}
%\vspace{-4mm}
\resizebox{\columnwidth}{!}
{
\begin{tabular}{|c|cccccc|}
\hline
Setting & 0+10 & 2+8 & 4+6 & 6+4 & 8+2 & 10+0 \\
\hline
Vimeo90K-Fast&36.10&\textcolor{blue}{36.13}& \textcolor{red}{36.18}&36.12& 36.04 &36.00   \\
\hline
Vimeo90K-Medium&34.86&\textcolor{blue}{34.90}& \textcolor{red}{34.96}& 34.88& 34.82 &34.79   \\
\hline
Vimeo90K-Slow&32.94 &\textcolor{blue}{32.98}&\textcolor{red}{33.02}& 32.96&32.92&32.88   \\
\hline
\end{tabular}
}
%\vspace{-5mm}
\label{frb ablation}
\end{table}

\begin{table}[t]
\smallskip
\centering
\setlength\tabcolsep{8pt}
\caption{PSNR (dB) evaluated by ablating DFU and FRU.}
%\vspace{-4mm}
\resizebox{\columnwidth}{!}
{
\begin{tabular}{|c|cccc|}
\hline
Setting & DFU & FRU & FRU+DFU & DFU+FRU \\
\hline
Vimeo90K-Fast& 35.88& 35.68& \textcolor{blue}{36.14}& \textcolor{red}{36.18}  \\
\hline
Vimeo90K-Medium& 34.62&34.63& \textcolor{blue}{34.89}& \textcolor{red}{34.96}   \\
\hline
Vimeo90K-Slow& 32.77& 32.79& \textcolor{blue}{32.98}& \textcolor{red}{33.02}  \\
\hline
\end{tabular}
}
%\vspace{-4mm}
\label{BCP ablation}
\end{table}

\heading{Different Numbers of HFB in HFM:} HFB is mainly to aggregate and distill structures and details components from temporal information to further refine spatial information. In this section, we conduct the ablation study to verify the impact of different numbers of HFB. As shown in Figure~\ref{cefm visual}, We can see that as the number of HFBs increases, the proposed network can generate more high-frequency information for texture restoration and clearer structure. Considering the trade-off between efficacy and efficiency, we set the numbers of HFB as 9 in the fusion process. The above experiments also show that more HFBs are helpful for spatial reconstruction by assigning learning specific tasks based on the divide-and-conquer strategy.

\section{Conclusions}
We proposed a YOGO method, which introduces a novel bidirectional interactive propagation module (BIPM) to sufficiently aggregate the spatial-temporal information. % Specifically, YOGO performs the backward and forward inferences in turn, where the hidden state of the backward inference is packed into the forward inference to super-resolve intermediate frames. In this way, 
Only one alignment and fusion are required in YOGO where feature representations can benefit from the past, current and future information via the bidirectional interaction. Furthermore, a Hybrid Fusion Module (HFM) is designed to aggregate and distill information to refine spatial information and reconstruct high-quality video frames.
Extensive quantitative and qualitative evaluations demonstrate our proposed method performs well against the state-of-the-art methods in ST-VSR tasks.

\heading{Acknowledgements}.
This work was supported by National Key R\&D Project (2021YFC3320301) and National Natural Science Foundation of China (62171325). The numerical calculations in this paper have been done on the supercomputing system in the Supercomputing Center of Wuhan University.

\bibliographystyle{ACM-Reference-Format}
\bibliography{sample-base}
\end{document}